\documentclass[lnbip]{svmultln}

\usepackage[ruled,vlined]{algorithm2e} 

\usepackage{graphicx}             	    
\usepackage{amsmath,amssymb}
\usepackage{multirow,multicol} 
\usepackage{listings}
\usepackage{color,soul}
\usepackage{url}

\usepackage{enumitem}
\setlist{topsep=0pt, leftmargin=*}

\newcommand{\figref}[1]{\mbox{Figure \ref{#1}}}
\newcommand{\tabref}[1]{\mbox{Table \ref{#1}}}
\graphicspath{{figs/}}

\begin{document}
\title{Using BERT Encoding to Tackle the\\ Mad-lib Attack in SMS Spam Detection}
\titlerunning{Tackling Mad-lib Spam Attack using BERT} 
\author{Sergio Rojas--Galeano\inst{*}}
\institute{$^*$Universidad Distrital Francisco Jos\'{e} de Caldas, Bogot\'{a}, Colombia\\ 
 \email{srojas@udistrital.edu.co}\\ 
}

\maketitle 

\begin{abstract} 
One of the stratagems used to deceive spam filters is to substitute vocables with synonyms or similar words that turn the message unrecognisable by the detection algorithms. In this paper we investigate whether the recent development of language models sensitive to the semantics and context of words, such as Google's BERT, may be useful to overcome this adversarial attack (called ``\mbox{Mad-lib}'' as per the word substitution game). Using a dataset of 5572 SMS spam messages, we first established a baseline of detection performance using widely known document representation models (BoW and TFIDF) and the novel BERT model, coupled with a variety of classification algorithms (Decision Tree, kNN, SVM, Logistic Regression, Naive Bayes, Multilayer Perceptron). Then, we built a thesaurus of the vocabulary contained in these messages, and set up a Mad-lib attack experiment in which we modified each message of a held out subset of data (not used in the baseline experiment) with different rates of substitution of original words with synonyms from the thesaurus. Lastly, we evaluated the detection performance of the three representation models (BoW, TFIDF and BERT) coupled with the best classifier from the baseline experiment (SVM). We found that the classic models achieved a 94\% \textit{Balanced Accuracy} (BA) in the original dataset, whereas the BERT model obtained 96\%. On the other hand, the Mad-lib attack experiment showed that BERT encodings manage to maintain a similar BA performance of 96\% with an average substitution rate of 1.82 words per message, and 95\% with 3.34 words substituted per message. In contrast, the BA performance of the BoW and TFIDF encoders dropped to chance. These results hint at the potential advantage of BERT models to combat these type of ingenious attacks, offsetting to some extent for the inappropriate use of semantic relationships in language. 
\end{abstract}


\section{Introduction}

Unsolicited email (spam) remains a global burden, accounting for up to 85\% of daily message traffic, according to some network security providers\footnote{see \url{https://dataprot.net/statistics/spam-statistics/}, last visit: July 13, 2021}. Although spam filters have taken advantage of artificial intelligence technologies to improve their detection performance, these algorithms can still be fooled by adversarial attacks, that is, carefully crafted content modifications that attempt to bypass the filters while nonetheless being easily recognisable to humans, by inoculating either benign, unrelated or obfuscated words or characters \cite{kuchipudi2020adversarial, rojas2017obstructing}. One of these kind of stratagems (called the ``Mad-lib '' attack in reference to the game based on replacing words in a sentence to come up with crazy stories), consists of substituting spam-triggering terms with synonyms or similar words preventing the message to be recognised as junk by the filter; this is also known as the synonym replacement attack \cite{kuchipudi2020adversarial}.

In this context, we envision using recent advances in semantic and context sensitive language models developed for natural language processing (NLP) tasks to combat these types of attacks. One of them is the BERT model developed by Google \cite{devlin2018bert}, which demonstrated state-of-the-art performance on eleven NLP tasks. In essence, this model is capable of representing a short document (consisting of a sequence of up to 512 words) as a numerical vector embedded in a space of 768 positions, corresponding to a dense and distributed representation of the document's features. Unlike embedding representations of individual words (such as Word2Vec, GloVe, or FastText, see for example \cite{kowsari2019text}), BERT is a deep network model that incorporates internal blocks of attention mechanisms that encode sequence of words into vectors depending on its context \cite{devlin2018bert}, capturing lexical, semantic and grammatical features related to the order in which, generally, one word precedes or succeeds another in particular sentences. One of the outputs of BERT is a vector representation of the input document, which we shall use to find similarities (distances) between spam messages in the resulting embedding space, likewise the word embeddings mentioned above are used to map similar words to close locations.

On this account, in this work we intend to evaluate the usefulness of applying the BERT model for the recognition of spam text sequences that differ only in some lexical terms but that still retain their unwanted intention, thus contributing to the detection of attacks of the Mad-lib type.

\subsection{Related work}
Adversarial attack tactics typically involve carefully crafting the content of the input data to disrupt the expected behaviour of a prediction model \cite{laskov2010machine}. The study of adversarial environments attracted attention more than a decade ago, when incidentally, the vulnerabilities of spam filters confronted with this type of manipulation were uncovered \cite{biggio2018wild}. Since then, many adversarial attacks and defences have been described in a variety of applications such as online abusive comments and profanity detection \cite{hosseini2017deceiving, rodriguez2018shielding, rojas2017obstructing, sood2012profanity}, classification of medical images \cite{finlayson2019adversarial} or object identification in computer vision \cite{hosseini2017google, akhtar2018threat}, to name a few. 

In the case of text classification tasks, attacks are generally performed by corrupting features or distorting the content of the text sequence \cite{rojas2017obstructing}. More particularly, in the field of adversarial attacks on spam filters, several tricks have been characterised \cite{imam2019survey, kuchipudi2020adversarial, rojas2013revealing}: poisoning, injection of good words, obfuscation of spam words, change of labels and replacement of synonyms. Our study focuses on the latter by taking a proactive approach \cite{biggio2018wild}, that is, anticipating, modelling and countering the adversarial strategy. In this sense, our study takes a step forward by showing the feasibility of addressing Mad-lib's adversaries (our second set of experiments, see below), compared to the work of \cite{kuchipudi2020adversarial} where the attack was described but was not addressed.

Regarding the use of BERT encodings for extracting spam features (our first set of experiments, see below), a modified Transformer model was recently proposed to improve the detection performance of spam classifiers \cite{liu2021spam}. Other modified models derived from BERT have been proposed for the effective detection of malicious phishing emails \cite{lee2020catbert}, while BERT with increased functionality has also been applied to filter multilingual spam messages \cite{cao2020bilingual} and to block fake tweets COVID \cite{kar2020no}, with promising results.

\subsection{Contributions}
The contributions of this study are:

\begin{itemize}
\item We show that the BoW, TFIDF, and BERT encoders are able to extract effective functions to identify spam using widely-used classification algorithms, with BERT performing slightly better. This corroborates findings previously reported in the literature (e.g.\cite{almeida2011contributions}.

\item We describe an automatic adversarial procedure to carry out a Mad-lib attack on the chosen dataset.

\item We provide empirical evidence that BERT is able to resist Mad-lib attacks whereas BoW or TFiDF are vulnerable.

\end{itemize}

\section{Methods}
\subsection{Study roadmap}
\begin{figure}[t]
    \centering
    \includegraphics[scale=.45]{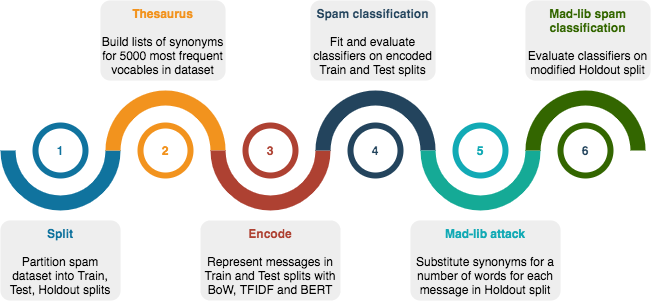}
    \caption{Study roadmap}
    \label{fig:roadmap}
\end{figure}

The study was conducted according to the stages illustrated in the roadmap of \figref{fig:roadmap}, which are described next.

    (1) \textit{Dataset splitting}. We worked with the SMS spam collection dataset from the UCI repository\footnote{\scriptsize The dataset is available at: \url{https://archive.ics.uci.edu/ml/datasets/sms+spam+collection}}. The dataset is unbalanced, as of the total of 5,574 messages, 4,827 are labelled as ham and only 747 as spam. The messages are quite short; with an average length of 14.5 words, they pose an interesting challenge for  content-based filtering algorithms \cite{almeida2011contributions}. We use random sampling without replacement to divide this data set into three subsets: train (60\%), test (20\%), and hold-out (20\%).

    (2) \textit{Thesaurus creation}. We extracted a vocabulary of the 5000 most frequent terms from the entire dataset and used them as keywords in a thesaurus. For each keyword, a list of synonyms was automatically scrapped from its corresponding entry page on the website \url{www.dictionary.com}. 
    
    (3) \textit{Document encoding}. Messages in each split are represented using two encodings commonly used in spam filtering, Bag-of-Words (BoW) and the Term Inverse Frequency of Document Frequency (TFIDF) \cite{kowsari2019text}, and the recently introduced Bi-directional Encoder Representations from Transformers (BERT) \cite{devlin2018bert}. BoW and TFIDF are simplified representations that map words within a document to a vector of frequencies indexed by a vocabulary (the latter normalised by the fraction of documents that contain the words). These mappings capture lexical features while ignoring syntax or semantics. For these models, we preprocess the text by removing accents, removing stopwords in English, converting it to lowercase, and applying stemming and tokenisation.
    
    On the other hand, BERT is a language model trained as a deep bidirectional network conditioned by both the left and right context of the words in the text input, also considering semantic relationships. One of the outputs at the top layer of the network is a vector of 768 positions that encodes an embedding of the entire input sentence. We will use it as a vector of context features and semantic relationships between a sequence of words that make up a message, focusing on its ability to project similar spam messages that differ in lexical variations in close locations of the embedding space, regardless of the actual interpretation of these features. Besides, the text cleanup for this model was minimal, basically converting to lowercase and applying the BERT tokeniser \cite{devlin2018bert}.
    
    (4) \textit{Spam classification}. At this stage, a first set of experiments was carried out to evaluate how well the classification algorithms work on the original messages. For this purpose, we used the training and test splits, represented with the three encodings as input features of a variety of classification algorithms that are regularly used for text classification tasks \cite{kowsari2019text,aggarwal2012survey,korde2012text}: Decision Tree, Naive Bayes, $k$-Nearest Neighbour (kNN), Support Vector Machine (SVM), Logistic Regression and Multilayer Perceptron (MLP). 
    
    (5) \textit{Mad-lib attack}.
    Two attacks were carried out on the held-out subset, where in each message an attempt was made to replace 5 or 10 words chosen at random, using synonyms from the previously constructed thesaurus. As a result, two modified Mad-lib subsets were obtained.
    
    (6) \textit{Mad-lib spam classification}. In this second set of experiments, the previously trained classifiers were evaluated in the modified Mad-lib sets, once encoded with the three aforementioned representation models.
    

\begin{figure}[b]
    \centering
    \includegraphics[scale=.43]{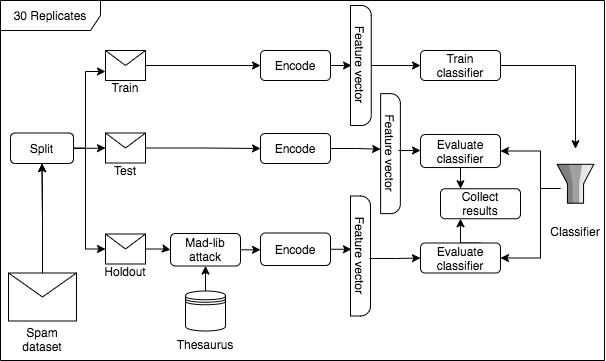}
    \caption{Experimental protocol}
    \vspace{-.6cm}
    \label{fig:protocol}
\end{figure}

\subsection{Experiment protocol}
The experiments were carried out according to the protocol described in \figref{fig:protocol}. The dataset is divided into three partitions: Train, Test, and Holdout. The first experiment was conducted to estimate a baseline of spam detection performance on the original data set, for comparison purposes in a subsequent spam experiment with the Mad-lib attack.

Initially, the messages in the Train and Test splits were encoded with the three representation models (BoW, TFIDF, BERT) to obtain vectors of 768 features (since this is the inherent size of the dense vectors generated by BERT, we set the vocabulary size base for BoW and TFIDF accordingly). Then the obtained feature vectors are fed to the aforementioned classification algorithms. Each classifier is trained with the encoded vectors of the Train split along with their respective labels; once trained, their performance is evaluated in the Test split, using the metrics of Accuracy (\textit{ACC}), Precision (\textit{PR}) and Sensitivity (\textit{SE}) \cite{tharwat2020classification} and Balanced Accuracy (\textit{BA}) \cite{brodersen2010balanced}. The latter was considered the most appropriate metric for this particular task, considering that the dataset is highly unbalanced. They are defined by the following equations:

{\footnotesize
\begin{align*}
ACC \!=\! \frac{TP+TN}{P+N},
\quad BA \!=\! \frac{1}{2}\left(\frac{TP}{P} + \frac{TN}{N}\right),
\quad 
PR \!=\! \frac{TP}{TP+FP},
\quad SE \!=\! \frac{TP}{TP+FN}, 
\end{align*}
}

\noindent where $P$ and $N$ are the total number of spam and ham messages, $TP$ and $FP$ are the correctly and wrongly classified spam, and $TN$ and $FN$ are the correctly and wrongly classified ham, respectively. The results are collected from a total of 30 replicas (with different samples of Train and Test partitions) so as to reduce their variability due to randomness in the sampling procedure.

The second set of experiments focused on evaluating how the previously trained classifiers perform on the Mad-lib attack (replacement of some words with synonyms) with the different representation models. To do this, the Hold-out split was used, together with the thesaurus to carry out two different attacks, one attempting to substitute 5 or 10 words at random. Notice that since some of the words may not have synonyms in the thesaurus, the actual number of substitutions may be less (see examples in \tabref{tab:examples}). Once the attack is completed, the modified messages are encoded with the three representation models to obtain the corresponding feature vectors, and they are fed to the previously trained classifiers, in order to evaluate their performance with the same metrics mentioned above. In this case, the results of the 30 replicas are averaged to reduce variability due to random sampling and word substitution processes.

\begin{table}[t]
\centering\footnotesize
\begin{tabular}{|p{11cm}|}
\hline
\underline{What} \underline{will} \underline{we} \underline{do} \underline{in} \underline{the} \underline{shower}, \underline{baby}?\\ 
what will we do in the \textbf{shower bath}  \textbf{infant}
\\ \hline

Good \underline{Morning} my \underline{Dear} \underline{..........}. Have a great \&\underline{amp}; \underline{successful} \underline{day}.\\ 
good \textbf{day} my \textbf{darling}            have a great  \textbf{ampere}  \textbf{victorious} \textbf{today}
\\ \hline\hline

\underline{Refused} \underline{a} loan\underline{? }  Secured \underline{or} Unsecured\underline{? }  Can'\underline{t}  \underline{get} credit? Call free now 0800 195 6669 \underline{or} text back '\underline{help}' \underline{\&} we will!\\ 
\textbf{turn down} a loan  secured or unsecured  can t \textbf{turn} credit  call free now 0800 195 6669 or text back  \textbf{care} we will
\\ \hline

\underline{Camera} \underline{-} \underline{You} are \underline{awarded} \underline{a} SiPix \underline{Digital} \underline{Camera} \underline{! } call 09061221066 \underline{fromm} landline. \underline{Delivery} within 28 days.\\ 
\textbf{cine-camera} you are \textbf{grant} a sipix \textbf{digital-analog converter} \textbf{cartridge}  call 09061221066 fromm landline \textbf{serving} within 28 days
\\ \hline
\end{tabular}
\caption{Examples of Mad-lib attacked ham (top) and spam (bottom) messages. Attempts: \underline{underline},  substitutions: \textbf{bold}. }
\label{tab:examples}
\end{table}

\subsection{Implementation details}
The models and experiments were implemented in the Python 3.8.5 language, using the libraries sckit-learn 0.24.0 \cite{pedregosa2011scikit}, PyDictionary \cite{pydictionary} and  SimpleTransformers \cite{simpletransformers}, which were executed in Google Colab with GPU accelerator. A repository with code and materials is available at: \url{github.com/Sargaleano/Madlib-Spam-Attack-BERT}.




\section{Results}

\subsection{Spam detection experiments}
The results for these experiments are summarised in \tabref{tab:classification}, where averages and standard deviations for the performance metrics are reported, grouped by encoding model and classification algorithm. A preliminary experimentation was conducted to perform classifier calibration (the final set of parameters is reported in the Appendix). 

\begin{table}[t]
\centering\footnotesize
\renewcommand{\arraystretch}{1.2}
\setlength{\tabcolsep}{5pt}
\begin{tabular}{|c|l|r|r|r|l|}
\hline
\textbf{Encoder}                & \multicolumn{1}{c|}{\textbf{Classifier}} & \multicolumn{1}{c|}{\textbf{BA}} & \multicolumn{1}{c|}{\textbf{ACC}} & \multicolumn{1}{c|}{\textbf{SE}} & \multicolumn{1}{c|}{\textbf{PR}} \\ \hline
\multirow{7}{*}{\textbf{BERT}}  & Decision Tree       & 85.2$\pm$1.7\%                      & 93.3$\pm$0.9\%                       & 73.9$\pm$3.1\%             & 76.6$\pm$4.1\%             \\ \cline{2-6} 
                                & Naive Bayes         & 93.1$\pm$1.0\%                      & 95.8$\pm$0.5\%                       & 89.3$\pm$2.0\%             & 82.0$\pm$3.2\%             \\ \cline{2-6} 
                                & kNN                 & 93.2$\pm$1.3\%                      & 97.0$\pm$0.5\%                       & 87.9$\pm$2.7\%             & 89.8$\pm$2.4\%             \\ \cline{2-6} 
                                & SVM (linear)        & 96.3$\pm$0.9\%                      & 98.4$\pm$0.3\%                       & 93.2$\pm$1.8\%             & 95.3$\pm$1.9\%             \\ \cline{2-6} 
                                & Logistic Regression & 96.3$\pm$0.9\%                      & 98.7$\pm$0.3\%                       & 93.0$\pm$1.8\%             & 97.5$\pm$1.3\%             \\ \cline{2-6} 
                                & MLP                 & 96.6$\pm$0.9\%                      & 98.8$\pm$0.3\%                       & 93.5$\pm$1.8\%             & 97.4$\pm$1.3\%             \\ \cline{2-6} 
                                & SVM (gaussian)      & 95.1$\pm$1.1\%                      & 98.6$\pm$0.4\%                       & 90.3$\pm$2.2\%             & 99.4$\pm$0.9\%             \\ \hline
\multirow{7}{*}{\textbf{BoW}}   & Decision Tree       & 84.1$\pm$1.8\%                      & 95.0$\pm$0.5\%                       & 69.3$\pm$3.8\%             & 91.3$\pm$3.2\%             \\ \cline{2-6} 
                                & Naive Bayes         & 82.6$\pm$1.2\%                      & 76.4$\pm$1.5\%                       & 91.0$\pm$2.2\%             & 35.1$\pm$2.7\%             \\ \cline{2-6} 
                                & kNN                 & 59.8$\pm$1.7\%                      & 89.3$\pm$0.8\%                       & 19.7$\pm$3.4\%             & 99.5$\pm$1.7\%             \\ \cline{2-6} 
                                & SVM (linear)        & 93.8$\pm$1.2\%                      & 97.7$\pm$0.3\%                       & 88.6$\pm$2.5\%             & 93.3$\pm$1.5\%             \\ \cline{2-6} 
                                & Logistic Regression & 92.8$\pm$1.4\%                      & 97.9$\pm$0.4\%                       & 85.9$\pm$2.7\%             & 97.8$\pm$1.4\%             \\ \cline{2-6} 
                                & MLP                 & 92.2$\pm$1.3\%                      & 97.7$\pm$0.4\%                       & 84.7$\pm$2.6\%             & 97.8$\pm$1.5\%             \\ \cline{2-6} 
                                & SVM (gaussian)      & 93.6$\pm$1.4\%                      & 97.5$\pm$0.4\%                       & 88.3$\pm$2.8\%             & 92.6$\pm$2.0\%             \\ \hline
\multirow{7}{*}{\textbf{TFIDF}} & Decision Tree       & 86.0$\pm$2.0\%                      & 94.9$\pm$0.6\%                       & 73.8$\pm$4.3\%             & 86.1$\pm$3.6\%             \\ \cline{2-6} 
                                & Naive Bayes         & 82.8$\pm$1.0\%                      & 77.9$\pm$1.5\%                       & 89.6$\pm$2.5\%             & 36.4$\pm$2.4\%             \\ \cline{2-6} 
                                & kNN                 & 55.4$\pm$1.1\%                      & 88.2$\pm$1.0\%                       & 10.8$\pm$2.2\%             & 99.4$\pm$1.9\%             \\ \cline{2-6} 
                                & SVM (linear)        & 94.0$\pm$1.4\%                      & 98.1$\pm$0.5\%                       & 88.5$\pm$2.8\%             & 96.6$\pm$2.0\%             \\ \cline{2-6} 
                                & Logistic Regression & 87.3$\pm$1.6\%                      & 96.5$\pm$0.5\%                       & 74.9$\pm$3.1\%             & 98.0$\pm$1.4\%             \\ \cline{2-6} 
                                & MLP                 & 88.0$\pm$1.5\%                      & 96.6$\pm$0.5\%                       & 76.2$\pm$3.1\%             & 97.9$\pm$1.3\%             \\ \cline{2-6} 
                                & SVM (gaussian)      & 93.9$\pm$1.4\%                      & 98.0$\pm$0.4\%                       & 88.4$\pm$2.9\%             & 96.3$\pm$1.8\%             \\ \hline
\end{tabular}
\caption{Spam classification results}
\label{tab:classification}
\end{table}

Roughly speaking, the results varied widely depending on the classification algorithm. For the BoW and TFIDF encoders, the lowest rates were obtained by kNN (\textit{BA}: 59.8\% and 55.4\% resp.), and the highest by SVM (\textit{BA}: 93.8\% and 94\% resp.). In the case of BERT, the variability is less noticeable (all classifiers obtained a \textit{BA} greater than 93\% except Decision Tree with 85\%), with MLP being the best achieving a \textit{BA} of 96.6\%, followed by SVM with 96.3\%. 

By examining the \textit{ACC} and \textit{SE} rates we corroborate similar results obtained in previous studies (for example \cite {almeida2011contributions}). We remark that the performances obtained for \textit {SE} and \textit {PR} with the BERT representation are more uniform than those of the other two encoders.

\subsection{Mad-lib spam attack experiments}
The results for these experiments are summarised in \tabref{tab:attack}, where averages and standard deviations for the performance metrics are reported, grouped by number of attempts in the attack and encoding model (linear SVM was chosen as classifier, since it achieved better performances across all of the three models).

In general, the results support the premise of the usefulness of the BERT model to resist this type of attack. We will focus on examining the \textit{BA} metric for this analysis. In the first attack with zero substitutions (that is, using the Holdout split without modifying the original messages) the SVM performance is maintained, with a value of 96.6\%. On the other hand, for the attacks with 5 and 10 substitution attempts (corresponding on average to 1.82 and 3.34 real substitutions as explained above), the accuracy rate of the BERT model decreased slightly to 96.2\% and 95.2\% respectively, about a 1\% drop compared to the baseline experiment.

In contrast, these results also show that regarding \textit{BA}, the performance of the BoW and TFIDF encoders degrades at levels close to chance. It is curious that even in the Hold-out partition without modifications the drop is noticeable; when examining the \textit{SE} rate, a sharp drop to 21.5\% is observed, that is, the detection of the features commonly associated to spam-related words is greatly affected by including out-of-sample terms, a phenomenon that is accentuated when Mad-lib substitutions are made in each message.

\begin{table}[t]
\footnotesize\centering
\renewcommand{\arraystretch}{1.2}
\setlength{\tabcolsep}{5pt}
\begin{tabular}{|c|c|r|r|r|r|r|}
\hline
\textbf{Attempts}   & \textbf{Encoder} & \multicolumn{1}{c|}{\textbf{Subs}} & \multicolumn{1}{c|}{\textbf{BA}} & \multicolumn{1}{c|}{\textbf{ACC}} & \multicolumn{1}{c|}{\textbf{SE}} & \multicolumn{1}{c|}{\textbf{PR}} \\ \hline
\multirow{3}{*}{0}  & BERT             & 0.00                               & 96.6$\pm$0.9\%                      & 98.3$\pm$0.4\%                       & 94.4$\pm$1.9\%                      & 92.8$\pm$2.3\%                      \\ \cline{2-7} 
                    & BoW              & 0.00                               & 54.9$\pm$3.8\%                      & 79.4$\pm$2.5\%                       & 21.5$\pm$8.0\%                      & 21.7$\pm$6.5\%                      \\ \cline{2-7} 
                    & TFiDF            & 0.00                               & 50.0$\pm$0.3\%                      & 86.6$\pm$0.8\%                       & 0.3$\pm$0.6\%                       & 13.7$\pm$28.4\%                     \\ \hline
\multirow{3}{*}{5}  & BERT             & 1.82                               & 96.2$\pm$1.0\%                      & 97.6$\pm$0.5\%                       & 94.2$\pm$2.0\%                      & 88.4$\pm$3.1\%                      \\ \cline{2-7} 
                    & BoW              & 1.82                               & 55.2$\pm$3.7\%                      & 80.7$\pm$2.2\%                       & 20.4$\pm$7.7\%                      & 23.4$\pm$7.0\%                      \\ \cline{2-7} 
                    & TFiDF            & 1.82                               & 50.0$\pm$0.3\%                      & 86.6$\pm$0.7\%                       & 0.3$\pm$0.6\%                       & 15.4$\pm$29.1\%                     \\ \hline
\multirow{3}{*}{10} & BERT             & 3.34                               & 95.2$\pm$0.9\%                      & 96.8$\pm$0.6\%                       & 93.0$\pm$1.7\%                      & 84.8$\pm$2.8\%                      \\ \cline{2-7} 
                    & BoW              & 3.34                               & 55.2$\pm$3.3\%                      & 82.1$\pm$2.1\%                       & 18.7$\pm$6.8\%                      & 25.8$\pm$7.8\%                      \\ \cline{2-7} 
                    & TFiDF            & 3.34                               & 50.0$\pm$0.2\%                      & 86.6$\pm$0.7\%                       & 0.2$\pm$0.4\%                       & 13.8$\pm$30.2\%                     \\ \hline
\end{tabular}
\caption{Mad-lib spam attack results}
\label{tab:attack}
\end{table}

\section{Conclusion}
This study provided empirical evidence on the promise of BERT encodings in tackling the Mad-lib spam attack. We reason that this is due to the ability of this model to represent semantic and contextual functions of language. Furthermore, other advantages of BERT are that it requires little pre-processing (cleanup) of text, as well as its ability to recognise out-of-vocabulary terms due to its inherent tokenisation method. On the computational side, BERT is heavier than the simpler BoW encoders that achieve comparable performances with spam not tampered with by Mad-lib adversaries.

Therefore, we anticipate that a combination of encoding models would be a realistic configuration at the core of modern spam filters, in order to detect behavioural changes implying that filter retraining is required (for example, activating an alert when the performance of BoW and BERT begins to differ amply). Furthermore, we hope that BERT encodings will help resist not only the adversarial scenario described in this document, but also other related attacks, such as the inoculation of good words, the obfuscation with homoglyphs, or the disguise of multilingual words that trigger spam. We plan to explore these ideas in our future work.


\bibliographystyle{plain}
\bibliography{biblio}

\section*{Appendix}
Chosen model parameters for algorithms used in experiments are shown below.

\vspace{.5cm}
\centering\footnotesize
\renewcommand{\arraystretch}{1.2}
\setlength{\tabcolsep}{5pt}
\begin{tabular}{|ll|}
\hline
\multicolumn{2}{|c|}{Classification algorithms} \\ \hline
\multicolumn{1}{|l|}{Decision Tree}        &  max\_depth=10 \\
\multicolumn{1}{|l|}{Naive Bayes}          &  default parameters\\
\multicolumn{1}{|l|}{kNN}                  &  k = 15 \\
\multicolumn{1}{|l|}{SVM (linear)}         &  C=1, loss=`squared\_hinge' \\
\multicolumn{1}{|l|}{Logistic Regression}  & default parameters \\
\multicolumn{1}{|l|}{MLP}                  &  hidden\_layer\_sizes=(10,), alpha=1, max\_iter=1000 \\
\multicolumn{1}{|l|}{SVM (gaussian)}       & gamma=.01, C=100 \\ \hline\hline
\multicolumn{2}{|c|}{Representation models}     \\ \hline
\multicolumn{1}{|l|}{BoW, TFIDF}           & stemming, lowercase, stop\_words, max\_features=768 \\
\multicolumn{1}{|l|}{BERT}                 & model=`xlm-r-bert-base-nli-stsb-mean-tokens' \\\hline
\end{tabular}

\end{document}